\documentclass[letterpaper, 10 pt, conference]{ieeeconf}  

\IEEEoverridecommandlockouts                              

\usepackage{graphics} 
\usepackage{epsfig} 
\usepackage{amsmath} 
\usepackage{amssymb}  
\usepackage{algorithm}
\usepackage{algpseudocode}
\usepackage{amssymb}
\usepackage{amsmath}
\usepackage{commath}
\usepackage{mathtools}

\usepackage{hyperref}
\usepackage{import}
\usepackage{paralist}
\usepackage{gensymb}
\usepackage{dsfont}
\usepackage{siunitx}
\usepackage[bottom]{footmisc}
\usepackage{framed}
\usepackage{balance} 
\usepackage{caption}
\usepackage[skins]{tcolorbox}  
\usepackage{subcaption}
\usepackage{bm}
\usepackage{cite} 
\usepackage{balance}

\hypersetup{
    colorlinks=true,
    linkcolor=black,
    citecolor=black,
    filecolor=black,
    urlcolor=black,
}
\DeclareCaptionFont{mysize}{\fontsize{8}{10}\selectfont}
\usepackage{soul}
\usepackage{hyperref}
\usepackage[percent]{overpic}
\usepackage{xcolor}
\newcommand{\insertWebPageLink}{\url{https://miladshafiee.github.io/ManyQuadrupeds/}}
\definecolor{ao(english)}{rgb}{0.0, 0.5, 0.0}
\definecolor{green(pigment)}{rgb}{0.0, 0.65, 0.31}

\usepackage{booktabs}

\usepackage[subtle,tracking=normal]{savetrees} 

\title{\hspace{-0.23cm}\Large \bf 
\textit{ManyQuadrupeds}:  Learning a Single Locomotion Policy for Diverse Quadruped Robots}

    \author{Milad Shafiee$^{1}$, Guillaume Bellegarda$^{1}$, and Auke Ijspeert$^{1}$ 
\thanks{$^{1}$ This research is supported by the Swiss National Science Foundation
(SNSF) as part of project No.197237. The authors are with the BioRobotics Laboratory, Ecole Polytechnique Federale de Lausanne (EPFL). {\tt\footnotesize (e-mail: firstname.lastname@epfl.ch)}}
}

\begin{document}
\bstctlcite{MyBSTcontrol}
\maketitle
\thispagestyle{empty}
\pagestyle{empty}

\begin{abstract}
Learning a locomotion policy for quadruped robots has traditionally been constrained to a specific robot morphology, mass, and size. The learning  process must usually be repeated for every new robot, where hyperparameters and reward function weights must be re-tuned to maximize performance for each new system.
Alternatively, attempting to train a single policy to accommodate different robot sizes, while maintaining the same degrees of freedom (DoF) and morphology, requires either complex learning frameworks, or  mass, inertia, and dimension
randomization, which leads to prolonged training periods.
In our study, we show that drawing inspiration from animal motor control allows us to effectively train a single locomotion policy capable of controlling a diverse range of quadruped robots. 
The robot differences encompass: a variable number of DoFs, (i.e. $12$ or $16$ joints), three distinct morphologies, a broad mass range spanning from $2$ \si{kg} to $200$ \si{kg}, and nominal standing heights ranging from $18$ \si{cm} to $100$ \si{cm}. 
Our policy modulates a representation of the Central Pattern Generator (CPG) in the spinal cord, effectively coordinating both frequencies and amplitudes of the CPG to produce rhythmic output (Rhythm Generation), which is then mapped to a Pattern Formation (PF) layer. 
Across different robots, the only varying component is the PF layer, which adjusts the scaling parameters for the stride height and length.  Subsequently, we evaluate the sim-to-real transfer by testing the single policy on both the Unitree Go1 and A1 robots. Remarkably, we observe robust performance, even when adding a 15 \si{kg} load, equivalent to $125\%$ of the A1 robot's nominal mass. 
\end{abstract}

\section{Introduction and Related Work}
\label{sec:INTRODUCTION}
The oldest group of vertebrates which have appendages (i.e. fins or legs) are the elasmobranchs (sharks and rays). These vertebrates have followed a separate evolutionary path from mammals for over 420 million years. However, neural circuits controlling elasmobranch fins and mammalian limbs have exhibited remarkable similarities at the molecular, cellular, and behavioral levels~\cite{grillner2020current}. This suggests that the neural substrate responsible for limb control was already present over 420 million years ago, and that the motor control scheme of tetrapods is preserved across various vertebrate species, each with their own unique size, inertia, and morphology~\cite{grillner2018evolution}. In contrast, in robotics, it is common practice to design and train a new control policy for each new specific robot. In this paper, we demonstrate how employing a biology-inspired motor-control scheme can streamline the training process, enabling the development of a single control policy applicable to quadruped robots with diverse sizes, inertias, morphologies, and degrees of freedom (DoF). 
\begin{figure}[h]
\centering
\includegraphics[scale=0.21038943783, trim ={0.0cm 0.0cm 0.0cm 0.0cm},clip]{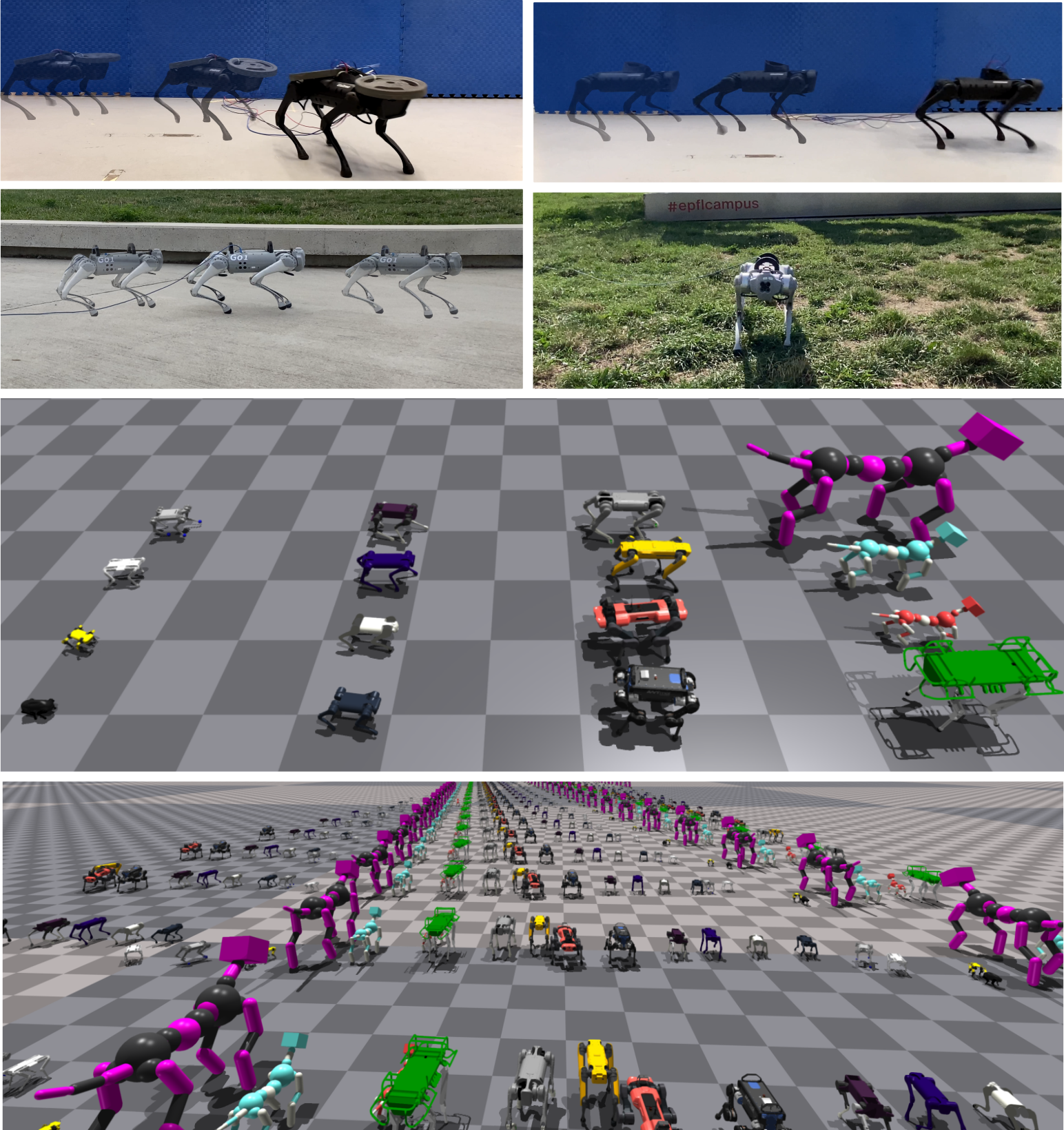}
\caption{
 Simulation and experiment snapshots of training and deploying a single policy to control 16 different robots. Videos: \insertWebPageLink }
\vspace{-2.0em}
     \label{fig:main}
\end{figure}
\subsection{Central Pattern Generators} 
Quadruped animal motor control can be described as an intricate interplay between the Central Pattern Generator (CPG, a system of coupled neural oscillators), sensory feedback, and supraspinal drive signals from the brain. In robotics, abstract models of CPGs are commonly used for locomotion pattern generation~\cite{mos2013cat,aoi2017adaptive,kimura2007adaptive,bellegardaquadFrog}, as well as to investigate biological hypotheses about animal motor control~\cite{ijspeert2007salamander, thandiackal2021emergence}. 
Besides the intrinsic oscillatory behavior of CPGs, several other properties such as robustness and implementation simplicity make CPGs desirable for locomotion control~\cite{ijspeert2008}. 


\subsection{Learning Locomotion}
Deep Reinforcement Learning (DRL) has emerged as a powerful approach for training robust legged locomotion control policies in simulation, and deploying these sim-to-real on hardware~\cite{peng2018deepmimic,iscen2018policies,hwangbo2019anymal,lee2020anymal,miki2022learning,siekmann2021blind,peng2020laikagoimitation,kumar2021rma,ji2022concurrent,margolis2022rapid,bellegarda2021robust,yang2022fast,shao2022gait,yu2023identifying}.  Most of these works view the trained artificial neural network (ANN) as a ``brain'' which has full authority  to directly control the joint movements.  These methods involve training custom policies from the ground up for each new quadruped's unique morphology. As a step towards generalization, Chiappa et al.~\cite{chiappa2022dmap} proposed an attention-based policy network architecture designed to train a policy capable of adapting to variations in body parameters.  Recent research aims to eliminate the need for policy retraining by investigating the potential of graph learning ~\cite{huang2020one,kurin2020my,trabucco2022anymorph,whitman2023learning}. {In these approaches, the agent's morphology is typically considered as a graph, with the graph's structure mirroring the agent's physical body.}
Most of these methods utilize agent-agnostic reinforcement learning combined with transfer learning techniques. 
While such methods based on graph learning exhibit promise in simulating toy examples, they have yet to be validated in real-world robotic systems~\cite{huang2020one,kurin2020my,trabucco2022anymorph}. Additionally, the gaits generated in these simulations often lack a natural sense of locomotion, which could  make the transition from simulation to the real world challenging. 


Feng et al.~\cite{feng2023genloco} introduced a unified policy training approach for learning locomotion through animal motion imitation for a diverse set of quadrupeds, encompassing a range of sizes and masses, yet sharing identical DoFs.
 This method demonstrated successful simulation-to-reality transfer capabilities; however, it requires a two-week training period on a 16-core CPU, and is limited to robots with the same number of DoFs per leg.
\subsection{Contribution}
In this paper, we employ a biology-inspired learning framework to learn a single locomotion policy able to control a diverse range of quadrupeds, encompassing significant variations in size, mass, morphology, and DoFs. Our proposed framework seamlessly integrates Central Pattern Generators and Deep Reinforcement Learning~\cite{bellegarda2022cpgrl,shafiee2023puppeteer,shafiee2023deeptransition,bellegarda2022visual}, where we use a simple Multi-Layer Perceptron (MLP) to represent higher control centers. This control center orchestrates the precise modulation of the CPG within the spinal cord and adeptly maps the rhythm generation network onto a pattern formation layer. We list below the advantages of the proposed architecture:

\textbf{Generality and Versatility:} The pattern formation layer, which shapes the feet trajectories in task-space, allows us to train a single locomotion policy for different robots with only a few parameter adjustments specific to each robot. Therefore, the action space does not rely on the number of joints or morphology, and we do not need to include joint information in the observation either. This leads to a constant size for both the action and observation spaces for all four-limb robots, which facilitates training robots with varying morphologies and DoFs with a simple MLP architecture. Mapping feet trajectories to joint positions is accomplished through inverse kinematics (IK). Solving the IK problem for legged robots, which fall under the category of serial manipulators, is a well-established research area. Numerous numerical and analytical methods are readily available for efficiently solving the IK problem for various types of serial limbs, each with different DoFs.
By entrusting the task of handling morphology to IK, in contrast to the current state-of-the-art  in learning quadrupedal locomotion for various robots~\cite{feng2023genloco}, our method  accommodates diverse morphologies and varying DoFs. Furthermore, we tested the generalization of the framework by excluding three robots with extreme mass and size properties across three different morphologies from the training process, but not for the testing phase. During these tests, we observed a stable gait for the previously unseen robots, even though the policy had not been specifically trained for them.

\textbf{Computational Efficiency:}
Additionally, our approach does not necessitate extensive dynamics randomization or motion imitation~\cite{feng2023genloco}, and it relies on a simple MLP architecture. This architectural simplicity, in addition to training robot policies in parallel on a single GPU in Isaac Gym~\cite{isaacgym}, results in high computational efficiency, enabling us to train a single policy for 16 diverse robots in less than two hours. 

\textbf{Robustness:}
Furthermore, we achieve stable trotting in sim-to-real quadruped experiments, even with an additional load of 15.0 \si{kg}, equivalent to $125\%$ of the A1 robot's nominal mass.  To the best of our knowledge, this accomplishment represents the pinnacle of robustness against additional loads ever achieved on the A1 robot~\cite{kumar2021rma,sombolestan2021adaptive,bellegarda2022cpgrl}.

\section{Central Pattern Generators}
\label{sec:cpg_intro}
\begin{figure}
\centering
\includegraphics[scale=0.813, trim ={0.0cm 0.cm 0.0cm 0.0cm},clip]{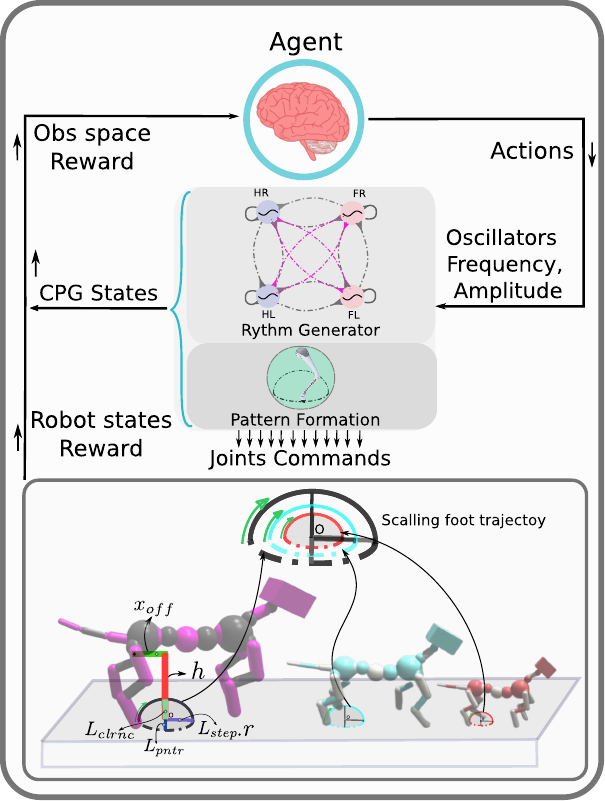}
\caption{
Quadrupedal locomotion in animals is governed by interactions between the spinal CPG, sensory feedback, and supraspinal brain signals. Here we employ DRL to train a neural network policy that mimics supraspinal drive behavior, enabling it to modulate CPG dynamics. 
To simulate the CPG within the spinal cord, we utilize nonlinear amplitude-controlled phase oscillators to represent the Rhythm Generator (RG) layer. The RG's outputs are then transformed into foot positions and, using inverse kinematics, converted into motor commands via a Pattern Formation (PF) layer. 
}
\label{fig:robot}
\vspace{-1.2em}
\end{figure} 
The vertebrate locomotor system is structured such that spinal CPGs are responsible for generating primary rhythmic patterns, while higher-level centers, such as the motor cortex, cerebellum, and basal ganglia, adjust these patterns in response to environmental conditions~\cite{grillner2020current}. Rybak et al.~\cite{rybak2006modelling}  propose that biological CPGs exhibit a dual-level functional organization, with a half-center rhythm generator (RG) determining locomotion frequency, and pattern formation (PF) circuits shaping the precise form of muscle activation signals~\cite{rybak2006modelling}. This two-tier functional model has also found application in robotics, particularly for quadruped locomotion research~\cite{bellegarda2022cpgrl,fukuhara2018spontaneous}.
\subsection{Rhythm Generator (RG) Layer}
We utilize non-linear phase oscillators to model the RG layer of the CPG circuits in the spinal cord. These oscillators have been effectively applied in the control of quadrupedal locomotion~\cite{sprowitz2013cheetah,ijspeert2007salamander,bellegarda2022cpgrl} with the following dynamics:
\begin{align}
\ddot{r}_i &= \alpha\left(\frac{\alpha}{4} \left(\mu_i - r_i \right) - \dot{r}_i \right) \label{eq:salamander_r} \\
\dot{\theta}_i &= \omega_i \label{eq:salamander_theta} 
\end{align}
where  $\theta_i$ is the phase of the oscillator, $r_i$ is the amplitude of the oscillator, $\omega_i$ and $\mu_i$  are the intrinsic frequency and amplitude, $\alpha$ is a positive convergence factor.
Notably, we do not consider explicit phase coupling between different oscillators. 
Consequently, the phase relationships between the legs must be learned and managed by the control policy by effectively modulating the intrinsic frequency for each limb. The control policy also learns to manipulate stride length by modifying the intrinsic amplitude.
\subsection{Pattern Formation (PF) Layer}
To establish a mapping from the output of the RG layer to joint commands, we first determine desired foot positions, and then we employ inverse kinematics to compute the corresponding desired joint positions. 
We formulate the desired foot position coordinates as follows:
\begin{align}
x_{i,\text{foot}} &= \ \ x_{off, i}  -L_{step} (r_i) \cos(\theta_i) \label{eq:feet-task1} \\
z_{i,\text{foot}} &= \begin{cases}
    z_{off, i}-h+ L_{clrnc}\sin(\theta_i) & \text{if } \sin(\theta_i) > 0 \\
    z_{off, i}-h+L_{pntr}\sin(\theta_i) & \text{otherwise}
\end{cases} 
\label{eq:feet-task}
\end{align}

\noindent where $L_{step}$ denotes the nominal step length, $h$ represents the scaling factor for body height, $L_{clrnc}$ indicates the maximum ground clearance during the swing, $L_{pntr}$ signifies the maximum ground penetration during stance phase, and $x_{off}$ is a set-point altering the equilibrium point of oscillation in the $x$ direction. The index $i$ corresponds to each limb.
$L_{step}$, $L_{clrnc}$, $L_{pntr}$, $h$, and $x_{off}$ are parameters that vary between different robots and are scaled based on the relative size of each robot, and these values remain constant throughout the training process.

\section{Learning Framework} 
\label{sec:RL}
In this section, we present our hierarchical bio-inspired learning framework for training a single policy to control various quadruped robots. We represent the supraspinal controller as an ANN, which we train with DRL so the agent can learn to modulate the intrinsic frequencies and amplitudes of each limb oscillation to produce stable gaits. We formulate the problem as a Markov Decision Process (MDP), which  consists of observations, actions, and rewards.  The proposed action-space modulates feet trajectories, and we do not include joint information in the observation space. This leads to a constant size for action and observation space for all four-limb robots, which facilitates training robots with various morphologies and DoFs with a simple MLP architecture. We detail the MDP components below. 
\subsection{Action Space}
\label{sec:action}
We incorporate one RG layer for each limb, as defined by Equations~(\ref{eq:salamander_r}) and~(\ref{eq:salamander_theta}). The RG output is then utilized in a PF layer to generate spatio-temporal foot trajectories in task space, as described in Equations~(\ref{eq:feet-task1}) and~(\ref{eq:feet-task}). Notably, we do not include explicit neural coupling, with the
intuition that inter-limb coordination will be controlled by the supraspinal drive.
As in~\cite{bellegarda2022cpgrl,shafiee2023deeptransition}, our action space modulates the intrinsic amplitudes and frequencies of the CPG, by continuously updating $\mu_i$ and $\omega_i$ for each leg. Therefore, our action space is represented as $\mathbf{a} = [\bm{\mu}, \bm{\omega}] \in \mathbb{R}^8$. The agent selects these parameters at a rate of 100 Hz, varying them at each step based on sensory inputs. During training, we impose the following limits on each input: $\mu \in [0.5, 4]$, and $\omega \in [0, 5]$ Hz. We emphasize that our action space modulates the feet trajectories in task-space, and these trajectories are subsequently mapped to joint space through inverse kinematics. Therefore, the proposed learning architecture is independent of the robot's morphology and DoFs, as long as we solve the inverse kinematics separately for each robot.
\begin{figure*}[h]
\centering
\includegraphics[scale=0.83, trim ={0.0cm 0.cm 0.0cm 0.0cm},clip]{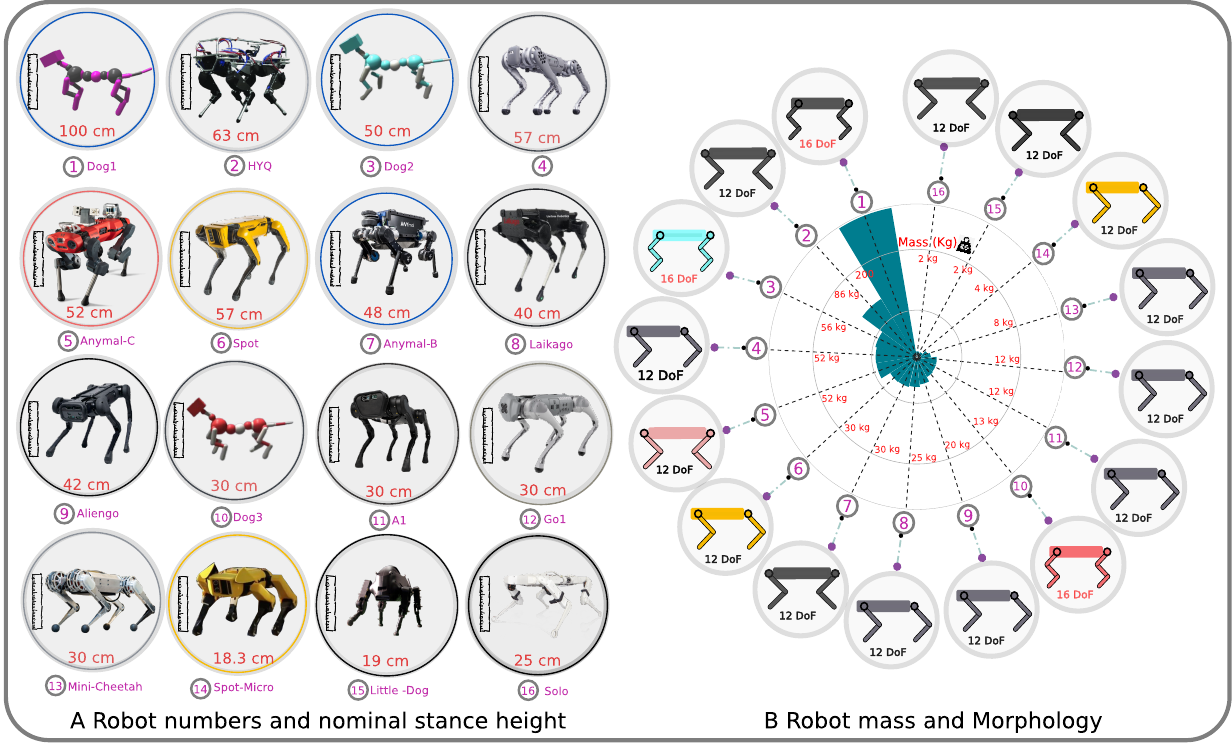}
\caption{Characteristics and parameters of 16 diverse quadruped robots. These robots exhibit variations in mass, ranging from 2 to 200 kg, nominal
height from 18 to 100 cm, and come in three different
morphologies, with two types of DoFs (either 12 or 16). }
     \label{fig:robot}
     \vspace{-1.5em}
\end{figure*} 
\subsection{Observation Space}
\label{sec:obs_space}
Our observation space includes body orientation, body linear and angular velocity, foot contact booleans, relative feet positions with respect to the body frame, the preceding action selected by the policy network, and the CPG states $\{\bm{r, \dot{r}, \theta, \dot{\theta}}\}$. It is worth noting that, unlike most other RL approaches, we omit proprioceptive information like joint positions and velocities. Instead, we employ a forward kinematics approach based on the current joint positions to determine relative foot positions with respect to the body frame, and then we incorporate these foot positions into the observation space. This design choice simplifies the training process and ensures it remains independent of specific morphologies and DoFs.

\subsection{Reward Function} 
\label{sec:reward}
We use the following reward function to promote viability with forward progress, minimize changes in body orientation, and encourage energy efficiency: 
\begin{equation}
\begin{split}
    R = w_1\cdot{\min}({f}_x, {d}_{max})   + w_2\cdot||\bm{o}_{base} - \bm{o}_{zero} || \\+  w_3\cdot|\bm{\tau} \cdot (\dot{\bm{q}}_t-\dot{\bm{q}}_{t-1})| 
\end{split}
\nonumber
\end{equation}
\begin{itemize}
\item \textit{Viability in forward progress}: 
In the first term, ${f}_x$ represents the robot's forward progress in the world (along the $x$-direction). To prevent the exploitation of simulator dynamics and the attainment of unrealistic speeds, we constrain this term. The constraint is set to ensure that the robot is rewarded for moving forward with a certain maximum distance during each control cycle, where ${d}_{max}$ denotes this maximum distance ($w_1 = 8.0$).
\item \textit{Base orientation penalty}: The second  term penalizes non-zero body orientation ($w_2 = -0.25$). 
\item \textit{Power}: The third term penalizes power to encourage energy-efficient gaits, with $\bm{\tau}$ and $\dot{\bm{q}}$ representing joint torques and velocities, respectively ($w_3 = -0.00001$).

\end{itemize}

\section{Results}
\label{sec:results}

In this section, we present results from learning a single unified policy to control multiple diverse quadrupeds in both simulation and hardware experiments. Section~\ref{subsec:implementation} details the implementation settings for both simulations and experiments.
We then delve into the outcomes of training the policy in Section~\ref{subsec:simulation}. Section~\ref{subsec:experiment} focuses on sim-to-real hardware results. For clear visualizations of the experiments, we encourage the reader to watch the supplementary video.

\subsection{Implementation Setting}
 \label{subsec:implementation}
We use Isaac Gym~\cite{isaacgym,rudin2022anymalisaac} for training and simulating 16 different quadruped robots. These robots exhibit variations in mass, ranging from 2 to 200 kg, nominal height from 18 to 100 cm, and come in three different morphologies, with two types of DoF, either 12 or 16. The robots in our study include the commercial robots Unitree A1, Go1, Aliengo, Laikago, B1, Boston Dynamics Spot, ANYbotics ANYmal-B and ANYmal-C, MIT mini-cheetah, Little-Dog, Spot-Micro, Solo, and HYQ, as well as three customized three-segmented leg quadruped robots. Characteristics and parameters for each robot are detailed in Table \ref{tab:robots} and Figure \ref{fig:robot}. 
We consider the following three main morphologies in simulation:

\noindent \textbf{Elbow-Up configuration for all limbs (3-DoF)}: 
     This configuration, used in robots like Spot, MIT mini-Cheetah, A1, Go1, B1, Laikago, Aliengo, and Spot-micro, consists of a 2-segmented elbow-up for both front and hind legs. Each leg has three Degrees of Freedom (DoF): one for adduction-abduction and two for hip and knee flexion/extension. We solve the inverse kinematics analytically with constraints to an elbow-up configuration.
     
\noindent  \textbf{Elbow-Up for front and Elbow-Down for hind limbs (3-DoF)}: 
   Similarly to the first category, the front limbs maintain an elbow-up configuration, but the hind limbs adopt an elbow-down configuration for the knee. We utilized the same analytical inverse kinematics solution as the first category but applied an elbow-down setting for the hind limbs. This configuration is used in robots like ANYmal-B, ANYmal-C, Solo, HYQ, and Little-Dog.

\noindent \textbf{Quadrupedal Animal-like Configuration (4-DoF)}: 
     In this morphology, inspired by quadruped animals, each leg  has 3 segments and four DoFs: one for adduction-abduction and three for hip, knee, and foot flexion/extension. We also use an analytical inverse kinematics solution for this configuration, which is applied to three animal-like robots with varying sizes and mass properties.  For a visual reference, please see Figure \ref{fig:robot}.

\begin{table*}[t]
\begin{center}
\begin{minipage}{\textwidth}
\caption{Characteristics and parameters of diverse quadruped  robots, which are fixed during training. The bold values represent the highest and lowest mass and dimensions, as well as a morphology that is unconventional for legged robots, more closely resembling the morphology of quadruped animals. }\label{tab:robots}
\vspace{-0.2em}
\begin{tabular*}{\textwidth}{@{\extracolsep{\fill}}cccccccccc@{\extracolsep{\fill}}}
\toprule%
Robot &  Height $h$ [\si{cm}]   & $L_{step}$ [\si{cm}] & $L_{clrnc}$ [\si{cm}] & $L_{pntr}$ [\si{cm}]& $x_{offset} $ [\si{cm}] & DoF - Morphology  &  Mass [\si{kg}] & $K_p$& $K_d$ \\
\midrule
Little Dog &19.0   & 5.0 & 4.7 & 0.5 & 1.1 &12 - 2 &2.9 & 20.0 & 0.3
\\

Spot-Micro &  \textbf{18.3} & 5.0 & 3.7 &0.5  & 1.0  &12 - 1  & 4.8 & 20.0 & 0.3
 \\

Solo & 25.0 &  10.0 & 5.0 & 0.5  & 3.7 & 12 - 2& \textbf{2.5} & 20.0 & 0.3 \\
 
Mini-Cheetah   &  30.0  & 13.0 &  7.0 & 1.0 & 0.0  &  12 - 1 & 8.4 & 100.0 & 2.7 \\ 

A1   & 30.0  & 13.0  & 7.0 & 1.0 & 0.0 &12 - 1 & 12.0  & 100.0 & 2.7\\

Go1  & 30.0 & 13.0  & 7.0 & 1.0  & 0.0 & 12 - 1 & 12.0 & 100.0 & 2.7
 \\
 
Aliengo  & 42.0 & 16.0   & 7.0  & 1.0  & 0.0 &12 - 1  & 20.6& 100.0 & 2.7
 \\

 Laikago & 40.0 & 16.0 & 7.0 & 1.0  & 0.0 &  12 - 1 &25.0 & 100.0 & 2.7 \\
 
 Anymal-B   & 48.0 & 17.0  & 7.0 & 0.0  & 10.0 &12 - 2  &30.0 & 430.0 & 20.7
 \\ 

Anymal-C   & 52.0 & 18.0  & 7.0 & 1.0  & 12.0 &12 - 2  & 52.1 & 430.0 & 20.7\\
Spot  & 57.0 & 20.0  & 9.0 & 1.0  & 0.0  &12 - 1  & 30.0 & 430.0 & 20.7  \\

B1   & 57.0 & 18.0  & 9.0 & 1.0  & 0.0 & 12 - 1 & 52.7 & 430.0 & 20.7\\

HYQ  & 63.0 & 20.0  & 9.0 & 1.0  &8.7  & 12 - 2  & 86.7 & 430.0 & 20.7\\

Dog1   & 30.0 & 13.0 & 7.0 & 1.0  & 0.0 & \textbf{16 - 3} &13.8 & 100.0 & 2.7 \\
Dog2   & 57.0 & 18.0  & 7.0 & 1.0  & 0.0 & \textbf{16 - 3} & 56.0 & 200.0 & 10.7\\
Dog3   & \textbf{100.0} & 36.0  & 9.0 & 2.0  &0.0  & \textbf{16 - 3} & \textbf{200.0} & 1400.0 & 140.7\\
\bottomrule
\vspace{-2.53em}
\end{tabular*}
\end{minipage}
\end{center}
\end{table*}
\begin{figure*}[h]
\centering
\includegraphics[scale=0.87013, trim ={0.0cm 0.cm 0.0cm 0.0cm},clip]{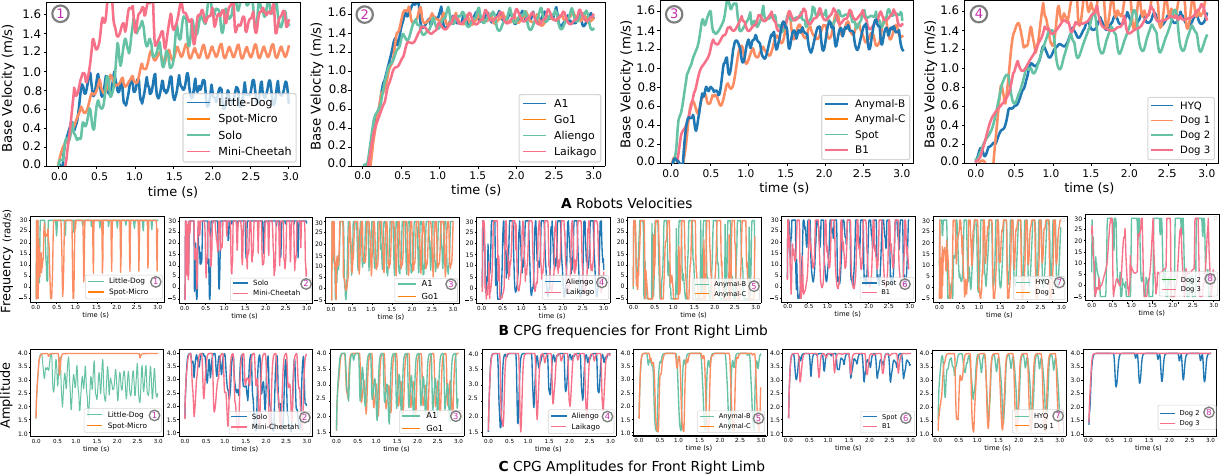}
\caption{Simulation results for training a single policy for 16 different quadrupeds: \textbf{Top:} Base locomotion velocity.   \textbf{Middle:} CPG Frequencies for Front Right limb.  \textbf{Bottom:} CPG Amplitudes for Front Right limb. 
{We use the FR limb since the robots exhibit a trot gait where the feet closely repeat the same pattern.}
}
\vspace{-1.5em}
     \label{fig:Policy1}
\end{figure*} 

To train the policies, we use Proximal Policy Optimization (PPO) \cite{ppo}, and Table \ref{table:RL} lists the PPO hyperparameters and neural network architecture.  The control frequency of the policy is 100 Hz, and the torques computed from the desired joint positions are updated at 1 kHz. The equations for each of the oscillators (Equations~\ref{eq:salamander_r} and~\ref{eq:salamander_theta}) are thus also integrated at 1 kHz.
All policies are trained for $14.0 \times 10^7$ samples on a {NVIDIA GeForce RTX 3070 8GB}. 

\begin{table}
\vspace{0.6em}
\centering
\caption{PPO Hyperparameters and neural network architecture used with Isaac Gym.} 
\vspace{-1.53em}
\vspace{0.73em}
\begin{tabular}{ c   c | c c   }
\hline
Parameter & Value & Parameter & Value \\
\hline
Batch size & 98304  & GAE discount factor & 0.95  \\
\hline
Mini-bach size & 24576  & KL-divergence $kl^*$ & 0.01\\ 
\hline
Number of epochs & 5  & Learning rate  & adaptive\\
\hline
Clip range & 0.2 & NN  Layers & [512, 256, 128]\\
\hline
Entropy coefficient & 0.01  & Activation & elu\\
\hline
Discount factor & 0.99 & Framework & Torch\\
\hline
\end{tabular} \\
\label{table:RL}
\vspace{-1.3em}
\end{table}

\begin{figure*}[t]
\centering
\includegraphics[scale=0.38, trim ={0.0cm 0.cm 0.0cm 0.0cm},clip]{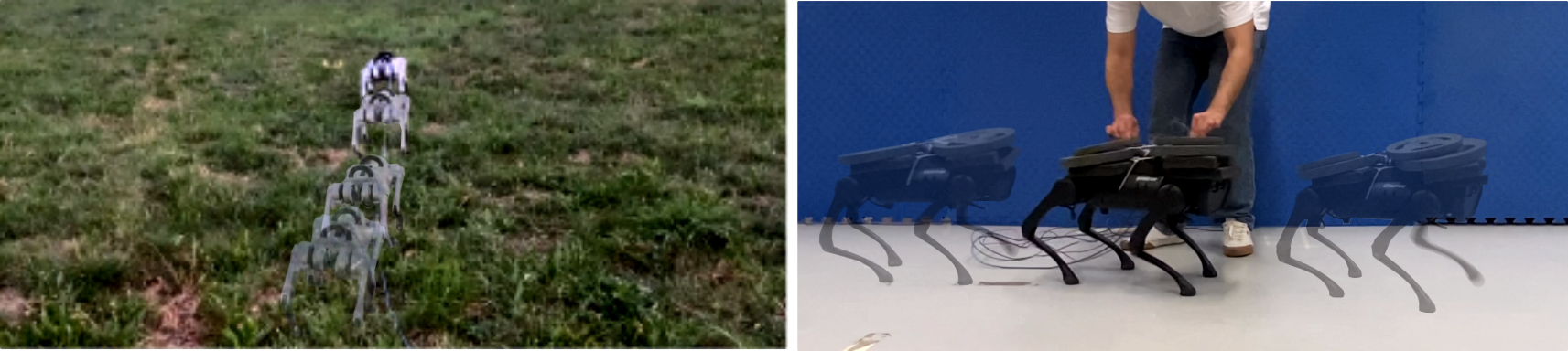}\\
\vspace{-0.2em}
\caption{ Left: Snapshots of Go1 trotting on uneven grass. Right: Snapshots of A1 carrying $10$ to $15$ \si{kg}. The robot starts trotting with a $10$ \si{kg} mass, and then an additional $5$ \si{kg} is added. The robot successfully walks despite never having encountered any such disturbances during training. 
}
     \label{fig:hw_snapshots}
     \vspace{-1.83em}
\end{figure*} 

\subsection{Simulation Results}
\label{subsec:simulation}
In this section, we present simulation results from training the 16 quadruped robots to learn forward locomotion. Four parameters need to be set for each robot, which are fixed during training: nominal standing height, nominal step length, ground clearance of feet during the swing phase, and feet penetration into the ground during stance phase, as shown in Table \ref{tab:robots}. 
These values are {heuristically } scaled for each robot based on its height in the zero joints position (i.e. when the legs are fully extended). 
Figure \ref{fig:Policy1} illustrates the base velocity and CPG frequency and amplitude {of the Front Right limb }  for this first trained policy across all quadruped robots. In this scenario, the reward function encourages forward movement as much as possible while penalizing velocities exceeding 1.5 \si{m/s}. Small-sized robots, such as Little-Dog and Spot-Micro, can only reach speeds of 0.8 and 1.2 \si{m/s}, respectively (Figure~\ref{fig:Policy1}-A-1). The frequency trajectory corresponding to small robots (Figure~\ref{fig:Policy1}-B-1 and 2)  shows that, in comparison to the larger robots (Figure~\ref{fig:Policy1}-B-3, 4, 5, 6), the policy tends to use the highest possible frequency to increase the velocity of small-sized robots. However, increasing velocity for the larger robots does not necessarily require very high frequencies, as they have greater flexibility to increase their base velocity by extending their stride length compared to smaller robots.

Our results show that robots with the second type of configuration such as Little-Dog, ANYmal-B, and ANYmal-C reach smaller velocities (0.8, 1.4, and 1.4 \si{m/s}) respectively) with respect to similar sized robots with the first type of morphology such as Spot-Micro, B1, and Spot (1.2, 1.55 and 1.55 \si{m/s} respectively). 
Furthermore, Figure~\ref{fig:Policy1}-C-5 and 6 show that ANYmal-B and C locomote with a smaller average amplitude, while Spot and Unitree B1 tend to have the highest amplitude possible, resulting in a longer stride length. Despite the similar mass properties of the compared robots, this observation needs a proper investigation for different nominal parameters to shed some light on the effect of morphology and mass properties in agility and efficiency, for future work.

An interesting observation is that Dog3, the largest simulated robot with a mass of $200$ \si{kg} and a nominal height of $100$ \si{cm}, utilizes the highest amplitude. In contrast to other robots, the amplitude reaches its maximum at the starting point of the movement and does not change, allowing for the use of the longest possible stride length, while maintaining a low locomotion frequency.
This observation intuitively suggests that robots with larger and heavier limbs tend to adopt a gait characterized by a low frequency and the longest possible stride length, which helps minimize the CoT (penalized in our reward function). At the other extreme, Spot-Micro, which is the smallest robot, exhibits a similar strategy for amplitude. It utilizes the highest possible amplitude, but unlike large robots, it takes many steps with a high control frequency. This behavior is driven by Spot-Micro's priority to increase speed due to its limitations caused by its small limbs. In contrast, increasing speed for Dog3 is less critical, so it selects a gait that minimizes energy consumption.

{Furthermore, we trained a single policy for 13 quadruped robots, excluding HYQ, Dog3, and B1 during training, and used these robots only for testing. We deliberately selected these robots to represent the extreme mass and size properties of each morphology to test the generalization capabilities of the framework. We observed reasonable locomotion behavior for these robots, even though the policy had not been specifically trained on them. Please refer to the supplementary video for visual reference.}
\subsection{Experimental Results}
 \label{subsec:experiment}
 For the hardware experiments, we trained the single policy to locomote with a maximum velocity of 1 \si{m/s}, without incorporating any domain randomization or noise during the training process. We transferred the trained policy sim-to-real to the Go1 and A1 robots, which are the only quadruped robots available in the lab. Figures~\ref{fig:main} and \ref{fig:hw_snapshots} show both simulation and experiment snapshots. 
We tested the control policy on the Go1 robot in two outdoor environments: one on a concrete pavement, and the other on uneven grass. In both scenarios, we observed a very smooth trotting gait. Notably, the grass surface is quite uneven, with holes and bumps that were not encountered during the training process.

We tested the same policy on the A1 robot in two types of experiments: normal walking, and load-carrying scenarios. It is noteworthy that the addition of loads was not encountered during training. We achieved stable trotting in all experiments, even with an additional load of 15.0 \si{kg}, equivalent to $125\% $ of the robot's nominal mass. To the best of our knowledge, this accomplishment represents the highest level of robustness against additional loads ever achieved with the A1 robot.

\section{Conclusion}
\label{sec:CONCLUSION}
Biological studies have shown that vertebrates  utilize very similar motor control architectures, despite large differences in morphology, size, mass, and DoFs~\cite{grillner2018evolution}.
Drawing inspiration from this fact, we have presented a biology-inspired learning framework based on Central Pattern Generators and Deep Reinforcement Learning (CPG-RL) with the capability to train a single policy for controlling diverse quadruped robots. These robots vary in the number of DoFs, with variations of $12$ and $16$, encompass three distinct morphologies, have a wide mass range of $2$ \si{kg} to $200$ \si{kg}, and nominal standing heights ranging from $18$ \si{cm} to $100$ \si{cm}. 
The proposed framework acts in task-space to modulate and control the feet trajectories, and it does not rely on observing joint information in the observation space, which facilitates training robots with different DoFs and morphologies. Moreover,
we are able to train a single policy that works for 16 different robots in less than two hours.
In our sim-to-real hardware experiments, we successfully demonstrated stable trotting even with an additional load of 15.0 \si{kg} on the Unitree A1 robot, which is equivalent to $125\%$ of the robot's nominal mass. To the best of our knowledge, this accomplishment demonstrates the highest level of robustness against additional loads ever achieved on the A1 robot.

For future work, we plan to expand our framework to include omni-directional motion planning on uneven terrain. 

\section*{Acknowledgements}
We would like to thank Alessandro Crespi for assisting with hardware setup. 

\bibliographystyle{IEEEtran}
\bibliography{bibliography}

\end{document}